\documentclass[11pt,a4paper]{article}
\usepackage[hyperref]{acl2020}
\usepackage{times}
\usepackage{latexsym}
\usepackage{graphicx}

\usepackage{tabu,multirow}
\usepackage{newunicodechar}
\newunicodechar{ℝ}{\mathbb{R}}

\usepackage{microtype}

\aclfinalcopy % Uncomment this line for the final submission
\title{A Report on the 2020 Sarcasm Detection Shared Task}

\author{Debanjan Ghosh\textsuperscript{1}, 
  Avijit Vajpayee\textsuperscript{1}, \textbf{and}
  Smaranda Muresan\textsuperscript{2}
   \\ 
  \textsuperscript{1} Educational Testing Service \\
  \textsuperscript{2}Data Science Institute, Columbia University \\
  {\tt \{dghosh, avajpayee\}@ets.org}\\
  {\tt smara@columbia.edu}}

\date{}

\begin{document}
\maketitle
\begin{abstract}
Detecting sarcasm and verbal irony is critical for understanding people’s actual sentiments and beliefs. Thus, the field of sarcasm analysis has become a popular research problem in natural language processing. As the community working on computational approaches for sarcasm detection is growing, it is imperative to conduct benchmarking studies to analyze the current state-of-the-art, facilitating  progress  in  this  area.  We report on  the  shared  task  on sarcasm detection we conducted as a part of the 2nd Workshop on Figurative Language Processing (FigLang 2020) at ACL 2020.
\end{abstract}

\section{Introduction} \label{section:intro}
Sarcasm and verbal irony are a type of figurative language where the speakers usually mean the opposite of what they say. Recognizing whether a speaker is ironic or sarcastic is essential to downstream applications for correctly understanding speakers' intended sentiments and beliefs. Consequently, in the last decade, the problem of irony and sarcasm detection has attracted a considerable interest from computational linguistics researchers. %\footnote{Although irony, sarcasm, and verbal irony  usually mean different expressions \cite{ghosh2018verbalirony}, for the sake of brevity we use them interchangeably in this report.}
%We notice the community working on this type of figurative language analysis is growing perhaps due to the abundance of figurative language content easily collected from social media platforms. 
%SM you said before sarcasm detection
%The greater proportion of NLP research on verbal irony or sarcasm has focused on the task of \emph{sarcasm detection} 
The task has been usually framed as a binary classification task (sarcastic vs. non-sarcastic) using either the utterance in isolation or adding contextual information such as conversation context, author context, visual context, or cognitive features \cite{davidov2010,tsur,gonzalez,riloff2013sarcasm,maynard2014cares,wallace2014humans,ghoshguomuresan2015,joshi2015harnessing,muresan2016identification,amir2016modelling,mishra2016harnessing,ghoshmagnet2017,felbo2017,ghoshsigdial2017,hazarika2018,tay2018,oprea2019exploring,majumder2019sentiment,castro2019towards,ghosh2019interpreting}. 

In this paper, we report on the shared task on sarcasm detection that we conducted as  part of the 2nd Workshop on Figurative Language Processing (FigLang
2020) at ACL 2020. The task aims to study the role of conversation context for sarcasm detection. Two types of social media content are used as training data for the two tracks - microblogging platform  such as Twitter and online discussion forum such as Reddit. %We only focus on detecting sarcasm that appears in  dialogue (in Twitter or Reddit).  

Table \ref{table:sarcrexamples_reddit} and Table \ref{table:sarcrexamples_twitter} show examples of three turn dialogues, where $Response$ is the sarcastic reply. % example of sarcastic turns responding to its prior turn $Context_{2}$ that in turn  a response to the turn $Context_{1}$. 
Without using the conversation context $Context_{i}$, it is difficult to identify the sarcastic intent expressed in $Response$.  The shared task is designed to benchmark the usefulness of modeling the entire conversation context  (i.e., all the prior dialogue turns) for sarcasm detection. %Note, for brevity, both examples contain only two prior turns as context, whereas, the number of prior turns is much more for many examples in the training corpus.
\begin{table}
\centering
\begin{tabular}{p{1.5cm}p{5.3cm}}
\hline
Turns & Message \\
\hline
$Context_{1}$ & The [govt] just confiscated a \$180 million boat shipment of cocaine from  drug traffickers.  \\
$Context_{2}$ & People think 5 tonnes is not a lot of cocaine.  \\
$Response$ & Man, I've seen more than that on a Friday night!  \\
\hline
\end{tabular}
\caption{Sarcastic replies to conversation context in Reddit. $Response$ turn is a reply to $Context_{2}$ turn that is a reply to $Context_{1}$ turn}
\label{table:sarcrexamples_reddit}
\end{table}

\begin{table}
\centering
\begin{tabular}{p{1.5cm}p{5.3cm}}
\hline
Turns & Message \\
\hline
$Context_{1}$ & This is the greatest video in the history of college football. \\
$Context_{2}$ & Hes gonna have a short career if he keeps smoking . Not good for your health \\
$Response$ & Awesome !!! Everybody does it. That’s the greatest reason to do something. \\
\hline
\end{tabular}
\caption{Sarcastic replies to conversation context in Twitter. $Response$ turn is a reply to $Context_{2}$ turn that is a reply to $Context_{1}$ turn}
\label{table:sarcrexamples_twitter}
\end{table}

 Section \ref{section:related} discusses the current state of research on sarcasm detection with a focus on the role of context. Section \ref{section:desc} provides a description of the shared task,  datasets,  and  metrics. Section \ref{section:systems} contains  brief  summaries   of each  of  the  participating  systems whereas Section \ref{section:results} reports  a  comparative evaluation  of  the  systems  and  our  observations about  trends  in  designs  and  performance  of  the systems that participated in the shared task.

\section{Related Work} \label{section:related}

% The main thing you want is utterance in isolation vs context. So I changed the orgnaization. Moved multi-modal part in ontext. 
A considerable amount of work on sarcasm detection has considered the utterance in isolation when predicting the sarcastic or non-sarcastic label. Initial approaches used feature-based machine learning models that rely on different types of features from lexical (e.g., sarcasm markers, word embeddings) to pragmatic such as emoticons or learned patterns of contrast between positive sentiment and negative situations \citep{davidov2010,veale2010detecting,gonzalez,liebrecht2013perfect,riloff2013sarcasm,maynard2014cares,joshi2015harnessing,ghoshguomuresan2015,ghosh2018marker}. Recently, deep learning methods have been applied for this task \cite{ghosh2016fracking,tay2018}.  For excellent surveys on sarcasm and irony detection see \cite{wallace2015computational,joshi2017automatic}. 

%With the recent advancement in leveraging the pre-trained language models, such as BERT for a variety of NLP problems    As discussed later in the paper, the fact that all but one of the participating teams for the shared task experimented with the transformer architectures testifies to the increasing popularity of this modeling approach. 

%SM is this shared task uttearance in isolation. 

However, when recognizing sarcastic intent even humans have difficulties sometimes when considering an utterance in isolation \cite{wallace2014humans}. Recently an increasing number of researchers have started to explore the role of contextual information for irony and sarcasm analysis. The term context loosely refers to any \textit{information} that is available beyond the utterance itself \cite{joshi2017automatic}. A few researchers have examined author context \cite{bamman2015contextualized,khattri2015your,rajadesingan2015sarcasm,amir2016modelling,ghoshmagnet2017}, multi-modal context \cite{schifanella2016detecting,cai2019multi,castro2019towards}, eye-tracking information \cite{mishra2016harnessing}, or conversation context \cite{bamman2015contextualized,wang2015twitter,joshi2016harnessing,zhang2016tweet,ghoshsigdial2017,ghoshmagnet2017}. 

Related to shared tasks on figurative language analysis, recently, \newcite{van2018semeval} have conducted a SemEval task on irony detection in Twitter focusing on utterances in isolation. Besides the binary classification task of identifying the ironic tweet the authors also conducted a multi-class irony classification to identify the specific \emph{type} of irony: whether it contains verbal irony, situational irony, or other types of irony. In our case, the current shared task aims to study the role of conversation context for sarcasm detection. In particular, we focus on benchmark the effectiveness of modeling the  conversation context (e.g., all the prior dialogue turns or a subset of the prior dialogue turns) for sarcasm detection.

%model the utterance and context separately to show that modeling conversation helps in sarcasm detection. Other types of context 

\section{Task Description} \label{section:desc}

%SM this is related work, I will move there. 
%Recently \cite{van2018semeval} have conducted a SemEval task on irony detection in Twitter. Besides the binary classification task of identifying the ironic tweet the authors also conducted a multi-class irony classification to identify the specific \emph{type} of irony: whether it contains verbal irony, situational irony, or other types of irony. 
%The Instead of looking at the type of irony or sarcasm, 
The design of our shared task is guided by two specific issues. First, we plan to leverage a particular type of context  --- the entire prior conversation context --- for sarcasm detection. Second, we plan to investigate the systems' performance on conversations from two types of social media platforms: Twitter and Reddit. Both of these platforms allow the writers to mark whether their messages are sarcastic (e.g., \#sarcasm hashtag in Twitter and ``/s'' marker in Reddit).  %

The competition is organized in two phases: training and evaluation. By  making available common datasets and frameworks for evaluation, we hope to contribute to the consolidation and strengthening of the growing  community  of  researchers  working  on  computational approaches to sarcasm analysis.

%SM You motivated the shared task need in intro, this is not design decision 
%Second, although there is a noticeable number of  publications on sarcasm detection, there is a lack of commonly used datasets. 
%SM the example you give all use different datasets... 
% except a few \cite{riloff2013sarcasm,ghoshguomuresan2015,ptavcek2014sarcasm,khodak2017large}. 

  % Recent research has focused on using \emph{conversation context}, but has been  \cite{bamman2015contextualized,ghosh2018clsarcasm}, however, they have not  not look into the role of the full dialogue threads as context. In this shared task, however, we provide training data with the full conversation context to detect sarcasm.

\subsection{Datasets} \label{subsection:data}

%SM I moved this above as it belongs to the task descriptoin and design choice. 
%One goal of the shared task is to comparatively study two types of social media platforms that have been considered individually for sarcasm detection:  discussion forums (e.g., Reddit) and Twitter. In the following sections, we briefly introduce the datasets used in the shared task.

\subsubsection{Reddit Training Dataset} 

\newcite{khodak2017large} introduced the self-annotated Reddit Corpus which is a very large collection of sarcastic and non-sarcastic posts (over one million) curated from different subreddits such as politics, religion, sports, technology, etc. This corpus contains self-labeled sarcastic posts where users label their posts as sarcastic by marking  ``/s'' to the end of sarcastic posts. For any such sarcastic post, the corpus also provides the full conversation context, i.e., all the prior turns that took place in the dialogue. 

We select the training data for the Reddit track from \newcite{khodak2017large}. We considered a couple of criteria. First, we choose sarcastic responses with at least two prior turns. Note, for many responses in our training corpus the number of turns is much more. Second, we curated sarcastic responses from a variety of subreddits such that no single subreddit (e.g., politics) dominates the training corpus. In addition, we  avoid responses from  subreddits that we believe are too specific and narrow (e.g.,  subreddit dedicated to a specific video game) that might not generalize well. The non-sarcastic partition of the training dataset is collected from the same set of subreddits that are used to collect sarcastic responses. We finally end up in selecting 4,400 posts (as well as their conversation context) for the training dataset equally balanced between sarcastic and non-sarcastic posts.

\subsubsection{Twitter Training Dataset} 
For the Twitter dataset, we have relied upon the annotations that users assign to their tweets using hashtags. The sarcastic tweets were collected using hashtags: \emph{\#sarcasm} and \emph{\#sarcastic}. As non-sarcastic utterances, we consider sentiment tweets, i.e., we adopt the methodology proposed in related work \cite{muresan2016identification}. Such sentiment tweets  do not contain the sarcasm hashtags but include hashtags that contain positive or negative sentiment words. The positive tweets express direct positive sentiment and they are collected based on tweets with positive hashtags such as \emph{\#happy}, \emph{\#love}, \emph{\#lucky}. Likewise, the negative tweets express direct negative sentiment and are collected based on tweets with negative hashtags such as  \emph{\#sad}, \emph{\#hate}, \emph{\#angry}. Classifying sarcastic utterances against sentiment utterances is a considerably harder task than classifying against random objective tweets since many sarcastic utterances also contain sentiment terms. Here, we are relying on \emph{self-labeled} tweets, thus, it is always possible that sarcastic tweets were mislabeled with sentiment hashtags or users did not use the \#sarcasm hashtag at all. We manually evaluated around 200 sentiment tweets and found very few such cases in the training corpus. Similar to the Reddit dataset we apply a couple of criteria while selecting the training dataset. First, we select sarcastic or non-sarcastic tweets only when they appear in a dialogue (i.e., begins with ``@''-user symbol) and at least have two or more prior turns as conversation context. Second, for the non-sarcastic posts, we maintain a strict upper limit (i.e., not-greater than 10\%) for any sentiment hashtag. Third, we apply heuristics such as avoiding short tweets, discarding tweets with only multiple URLs, etc. %Finally, we also manually remove instances that are thematically identical. A large number of recent sarcastic tweets are on politics and we removed some to keep a balance between different themes. 
We end up selecting 5,000 tweets for training balanced between sarcastic and non-sarcastic tweets.

\begin{figure}[t]
\centering

\includegraphics[width=7.5cm]{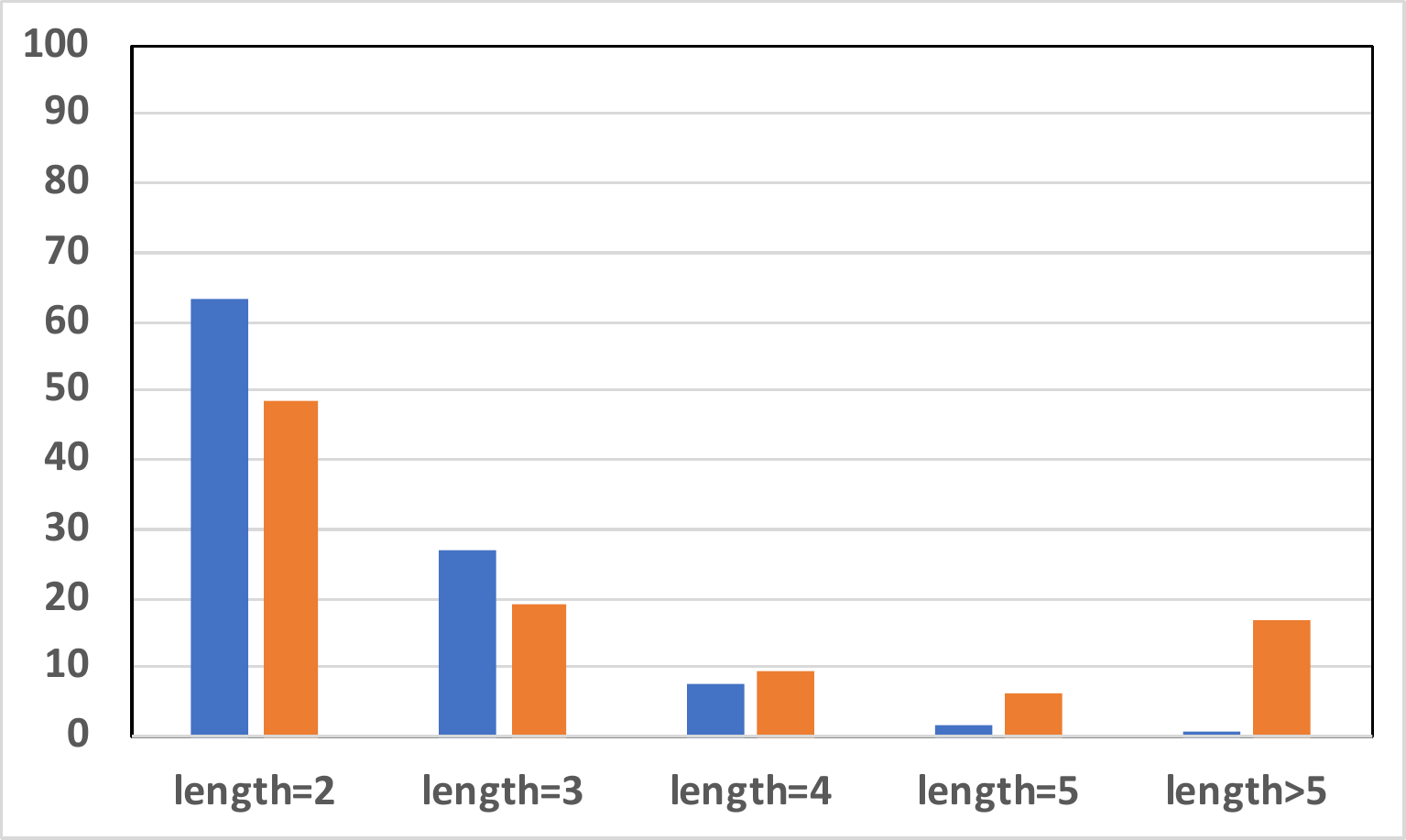}

%\end{framed}
\caption{Plot of Reddit (blue) and Twitter (orange) training datasets on the basis of context length. X-axis represents context length (i.e., number of prior turns) and Y-axis represents the \% of training utterances.}
\label{figure:histo}
\end{figure}

Figure \ref{figure:histo} presents a plot of number of training utterances on the basis of context length, for Reddit and Twitter tracks respectively. We notice, although the numbers are comparable for utterances with context length equal to two or three, for Twitter corpus, utterances with a higher number of context (i.e., prior turns) is much higher. 

\subsubsection{Evaluation Data}

The Twitter data for evaluation is curated similarly to the training data. For Reddit, we do not use \newcite{khodak2017large} rather collected new sarcastic and non-sarcastic responses from Reddit. %SM why?
First, for sarcastic responses we utilize the same set of subreddits utilized in the training dataset, thus, keeping the same genre between the evaluation and training. For the non-sarcastic partition, we utilized the same set of subreddits and submission threads as the sarcastic partition. For both tracks the evaluation dataset contains 1800 instances partitioned equally between the sarcastic and the non-sarcastic categories.    %SM 1800 in each class or 900 S and 900 NS?

\subsection{Training Phase}
In  the first  phase,   data  is  released  for  training  and/or  development of  sarcasm 
detection models (both Reddit and Twitter). Participants can choose to partition the training data further to a validation set for preliminary evaluations and/or tuning of hyper-parameters. Likewise, they can also elect to perform cross-validation on  the  training data. 

\subsection{Evaluation Phase}

In  the second phase,  instances  for  evaluation  are released. Each  participating  system  generated predictions  for  the  evaluation  instances,  for  up  to $N$ models. \footnote{$N$ is set to 999.} Predictions are submitted to the CodaLab site and evaluated   automatically   against   the   gold labels. CodaLab is an established platform to organize shared-tasks \cite{leong2018report} because it is easy to use, provides easy communication with the participants (e.g., allows mass-emailing) as well as tracks all the submissions updating the leader-board in real-time. The  metrics  used for  evaluation  is  the average F1 score between the two categories - sarcastic and non-sarcastic.  The leaderboards displayed the Precision, Recall, and F1 scores in the descending order of the F1 scores, separately for the two tracks - Twitter and Reddit. 
%SM make sure spelling is leaderboard not leader-board

\section{Systems} \label{section:systems}
\begin{table*}[ht]
\centering
\begin{tabular}{p{.5cm}p{.5cm}p{1.5cm}p{1cm}p{1cm}p{1cm}p{7.5cm}}
\hline
Rank & Lb. Rank & Team & P & R & F1 & Approach \\
\hline
1 & 1 & miroblog & 0.834 &	0.838 &	0.834 & BERT + BiLSTM + NeXtVLAD + Context Ensemble + Data Augmentation\\
2 & 2 & andy3223 & 0.751 &	0.755 &	0.750 &  RoBERTa-Large (all the  prior turns) \\
3 & 6 & taha & 0.738 &	0.739 &	0.737 & BERT+ Local Context Focus \\
4 & 8 &	tanvidadu &	0.716 &	0.718 &	0.716 &	RoBERTa-Large (last two prior turns) \\
5 & 9 &	nclabj	& 0.708	& 0.708	& 0.708	& RoBERTa + Multi-Initialization Ensemble \\
6 & 12	& ad6398 &	0.693 &	0.699 &	0.691 &	RoBERTa + LSTM \\
7 & 16 &	kalaivani.A	& 0.679	& 0.679	& 0.679	& BERT (isolated response) \\
8 & 17 & amitjena40	& 0.679 &	0.683 &	0.678 &	TorchMoji + ELMO + Simple Exp. Smoothing \\
9 & 21 &	burtenshaw	& 0.67	& 0.677	& 0.667	& Ensemble of SVM, LSTM, CNN-LSTM, MLP \\
10 & 26 & salokr & 0.641 & 0.643 & 0.639 & BERT + CNN + LSTM \\
11 & 31 & adithya604 & 0.605 & 0.607 & 0.603 & BERT (concatenation of prior turns and response) \\
12 & - & baseline & 0.600 & 0.599 &  0.600 &
$LSTM_{attn}$ \\
13 & 32 & abaruah & 0.595 &	0.605 &	0.585 & BERT-Large (concatenation of response and its immediate prior turn) \\

\hline
\end{tabular}
\caption{Performance of  the  best  system  per  team  and  baseline  for  the  Reddit track. We include two ranks - ranks from the submitted systems as well as the Leaderboard ranks from the CodaLab site}
\label{table:redditresults}
\end{table*}
%SM You have evaluation data and then test data: be consistent. I guess call it Test data? (you will need to fix previous section too)
The shared task started on January 19, 2020, when the training data was made available to all the  registered participants. We released the evaluation data on  February  25,  2020. Submissions were accepted until March 16, 2020. Overall, we received an overwhelming number of submissions: 655 for the Reddit track and 1070 for the Twitter track. The CodaLab leaderboard showcases results from 39 systems for the Reddit track and 38 systems for the Twitter track, respectively. %SM is this by setup why different numbers for different tracks? 
Out of all submissions, 14 shared task system  papers  were  submitted. %SM What are the reviews for this paper? We need more details what is wrong in the ppaer, it means this paper shouldn't be accepted as shared task paper. 
In the following section we  summarize each system paper. We also put forward a comparative analysis based on their performance and the choice of features/models in Section \ref{section:results}. Interested readers can refer to the individual teams' papers for more details. But first, we discuss the baseline classification model that we used.

\subsection{Baseline Classifier}

%SM Please spell out all acronyms first time they appear including LSTM, BERT etc throuhgut the paper. Once you introduce the acronym you can then use it. 
We use prior published work as the baseline that used conversation context to detect sarcasm from social media platforms such as Twitter and Reddit \cite{ghosh2018clsarcasm}. \newcite{ghosh2018clsarcasm} proposed a \emph{dual LSTM architecture} with hierarchical attention where one LSTM models the  conversation context and the other models sarcastic response.  The hierarchical attention \cite{yang2016hierarchical} implements two levels of attention -- one at the word level and another at the sentence level. We used their system based on only the immediate conversation context (i.e., the immediate prior turn). \footnote{\url{https://github.com/Alex-Fabbri/deep\_learning\_nlp\_sarcasm}}  This is denoted as $LSTM_{attn}$ in Table \ref{table:redditresults} and Table \ref{table:twitterresults}. 

\subsection{System Descriptions}

We describe the participating systems in the following section (in  alphabetical order).  
%SM Are you reporting the best performing system from each participant then put a sentence like
% The best-performing systems from each participant is described below in X order... 
%SM what is the order you are presenting them in? It is not rank, it is not alphabetical. It is not model... Fine to do alphabetical and then replace X above with alphabetical. 

%SM please spell out acronymis in all these desription if first introduced
\paragraph{abaruah \cite{abaruah2020}:} Fine-tuned a BERT model \cite{devlin2018bert} and reported results on varying maximum sequence length (corresponding to varying level of context inclusion from just response to entire context). They also reported results of BiLSTM with FastText embeddings (of response and entire context) and  SVM based on char n-gram features (again on both response and entire context). One interesting result was SVM with discrete features performed better than BiLSTM. They achieved best results with BERT on response and most immediate context.

\paragraph{ad6398 \cite{ad63982020}:} Report results comparing multiple transformer architectures (BERT, SpanBERT \cite{joshi2020spanbert}, RoBERTa \cite{liu2019roberta}) both in single sentence classification (with concatenated context and response string) and sentence pair classification (with context and response being separate inputs to a Siamese type architecture). Their best result was with using RoBERTa + LSTM model.

\paragraph{aditya604 \cite{aditya6042020}:} Used BERT on simple concatenation of last-k context texts and response text. The authors included details of data cleaning (de-emojification, hashtag text extraction, apostrophe expansion) as well experiments on other architectures (LSTM, CNN, XLNet \cite{yang2019xlnet}) and varying size of context (5, 7, complete) in their report. The best results were obtained by BERT with 7 length context for Twitter dataset and BERT with 5 context for Reddit dataset.

\paragraph{amitjena40 \cite{amitjena402020}:} Used a time-series analysis inspired approach for integrating context. Each text in conversational thread (context and response) was individually scored using BERT and Simple Exponential Smoothing (SES) was utilized to get probability of final response being sarcastic. They used the final response label as a pseudo-label for scoring the context entries, which is not theoretically grounded. If final response is sarcastic, the previous context dialogue cannot be assumed to be sarcastic (with respect to its preceding dialogue). However, the effect of this error is  attenuated due to exponentially decreasing contribution of context to final label under SES scheme. %SM what is SES, again acronym

\paragraph{AnandKumaR \cite{AnandKumaR2020}:} Experimented with using traditional ML classifiers like SVM and Logisitic Regression over embeddings through BERT and 
GloVe \cite{pennington2014glove}. Using BERT as a feature extraction method as opposed to fine-tuning it was not beneficial and Logisitic Regression over GloVe embeddings outperformed them in their experiment. Context was used in their best model but no details were available about the depth of context usage (full vs. immediate). Additionally, they only experimented with Twitter data and no submission was made to the Reddit track. They provided details of data cleaning measures for their experiments which involved stopword removal, lowercasing, stemming, punctuation removal and spelling normalization. 

\paragraph{andy3223 \cite{andy32232020}:} Used the transformer-based architecture for sarcasm detection, reporting the performance of three architecture, BERT, RoBERTa, and ALBERT \cite{lan2019albert}. They considered two models, the \emph{target-oriented} where only the target (i.e., sarcastic response) is modeled and \emph{context-aware}, where the context is also modeled with the target. The authors conducted extensive hyper-parameter search, and set the learning rate to 3e-5, the number of epochs to 30, and use different seed values, 21, 42, 63, for three runs. Additionally, they set the maximum sequence length 128 for the \emph{target-oriented} models while it is set to 256 for the \emph{context-aware} models.

\paragraph{burtenshaw \cite{burtenshaw2020}:} Employed an ensemble of four models - LSTM (on word, emoji and hashtag representations), CNN-LSTM (on GloVe embeddings with discrete punctuation and sentiment features), MLP (on sentence embeddings through Infersent \cite{conneau2017supervised}) and SVM (on character and stylometric features). The first three models (except SVM) used the last two immediate contexts along with the response.

\paragraph{duke\_DS \cite{dukeds2020}:} Here the authors have conducted extensive set of experiments using discrete features, DNNs, as well as transformer models, however, reporting only the results on the Twitter track. Regarding discrete features, one of novelties in their approach is including a \emph{predictor} to identify whether the tweet is political or not, since many sarcastic tweets are on political topics. Regarding the models, the best performing model is an ensemble of five transformers:  BERT-base-uncased, RoBERTa-base, XLNet-base-cased, RoBERTa-large, and ALBERT-base-v2.   

\paragraph{kalaivani.A \cite{kalaivani2020}:} Compared traditional machine learning classifiers (e.g., Logistic Regression/Random Forest/XGBoost/Linear SVC/ Gaussian Naive Bayes) on discrete bag-of-word features/Doc2Vec features with LSTM models on Word2Vec embeddings \cite{mikolov2013distributed} and BERT models. For context usage they report results on using isolated response, isolated context and context-response combined (unclear as to how deep the context usage is). The best performance for their experiments was by BERT on isolated response.

\paragraph{miroblog \cite{miroblog2020}:} Implemented a classifier composed of BERT followed by BiLSTM and NeXtVLAD \cite{lin2018nextvlad} (a differentiable pooling mechanism which empirically performed better than Mean/Max pooling). \footnote{VLAD is an acronym of ``Vector of Locally Aggregated Descriptors'' \cite{lin2018nextvlad}.}  They employed an ensembling approach for including varying length context and reported that gains in F1 after context of length three are negligible. Just with these two contributions alone, their model outperformed all others. Additionally, they devised a novel approach of data augmentation (i.e., Contextual Response Augmentation) from unlabelled conversational contexts based on next sentence prediction confidence score of BERT. Leveraging large-scale unlabelled conversation data from web, their model outperformed the second best system by 14\% and 8.4\% for Twitter and Reddit respectively (absolute F1 score).

\paragraph{nclabj \cite{nclabj2020}:} Used a majority-voting ensemble of RoBERTa models with different weight-initialization and different levels of context length. Their report shows that previous 3 turns of dialogues had the best performance in isolation. Additionally, the present results comparing other sentence embedding architectures like Universal Sentence Encoder \cite{cer2018universal}, ELMo \cite{peters2018deep} and BERT.

\paragraph{salokr/vaibhav \cite{salokr2020} :} Employed a CNN-LSTM based architecture on BERT embeddings to utilize the full context thread and the response. The entire context after encoding through BERT is passed through CNN and LSTM layers to get a representation of the context. Convolution and dense layers over this summarized context representation and BERT encoding of response make up the final classifier. 

%SM what is LCF? What is AEN again acronyms. it is not readable. 
\paragraph{taha \cite{taha2020}:} Reported experiments comparing SVM on character n-gram features, LSTM-CNN models, Transformer models as well as a novel usage of aspect based sentiment classification approaches like Interactive Attention Networks(IAN) \cite{ma2017interactive}, Local Context Focus(LCF)-BERT \cite{zeng2019lcf} and BERT-Attentional Encoder network (AEN) \cite{song2019attentional}. For aspect based approaches, they viewed the last dialogue of conversational context as aspect of the target response. LCF-BERT was their best model for the Twitter task but due to computational resource limitations they were not able to try it for Reddit task (where BERT on just the response text performed best).

\paragraph{tanvidadu \cite{tanvidadu2020}:} Fine-tuned RoBERTa-large model (355 Million parameters with over a 50K vocabulary size) on response and its two immediate contexts. They reported results on three different types of inputs: response-only model, concatenation of immediate two context with response, and using an explicit separator token between the response and the final context. The best result is reported in the setting where they used the separation token.

\begin{table*}[ht]
\centering
\begin{tabular}{p{.5cm}p{.5cm}p{1.5cm}p{1cm}p{1cm}p{1cm}p{7.5cm}}
\hline
Rank & Lb. Rank & Team & P & R & F1 & Approach \\
\hline
1 & 1 &	miroblog	& 0.932 &	0.936 &	0.931 &	BERT + BiLSTM + NeXtVLAD + Context Ensemble + Data Augmentation \\
2 & 2 &	nclabj&	0.792&	0.793&	0.791&	RoBERTa + Multi-Initialization Ensemble \\
3 & 3 & andy3223 & 0.791 &	0.794 &	0.790 & RoBERTa-Large (all the prior turns) \\
4 &5 &	ad6398 &	0.773 &	0.774 &	0.772 &	RoBERTa + LSTM\\
5 & 6 &	tanvidadu &	0.772 &	0.772 &	0.772 &	RoBERTa-Large (last two prior turns)\\
6 & 8 & duke\_DS & 0.758 &	0.767 &	0.756 & Ensemble of Transformers \\
7 & 11 & amitjena40 &	0.751 &	0.751 &	0.750 &	TorchMoji + ELMO + Simple Exp. Smoothing \\
8 & 13 & salokr & 0.742 & 0.746 & 0.741 & BERT + CNN + LSTM \\
9 & 16 &	burtenshaw &	0.741 &	0.746 &	0.740 &	Ensemble of SVM, LSTM, CNN-LSTM, MLP \\
10 & 21 & abaruah & 0.734 & 0.735 & 0.734 & BERT-Large (concatenation of response and its immediate prior turn) \\
11 & 24	& taha	& 0.731	& 0.732 &	0.731 &	BERT \\
12 & 27 & kalaivani.A & 0.722 & 0.722 & 0.722 & BERT (isolated response)  \\
13 & 28 & adithya604 &
0.719 &	0.721 &	0.719 & BERT (concatenation of prior turns and response) \\
14 & 35 & AnadKumR   &   0.690 & 0.690 & 0.690 & GloVe + Logistic Regression \\
15 & - & baseline & 0.700 & 0.669 &  0.680 & $LSTM_{attn}$ \\

\hline
\end{tabular}
\caption{Performance of  the  best  system  per  team  and  baseline  for  the  Twitter track. We include two ranks - ranks from the submitted systems as well as the Leaderboard ranks from the  CodaLab site}
\label{table:twitterresults}
\end{table*}

\section{Results and Discussions} \label{section:results}

Table \ref{table:redditresults} and Table \ref{table:twitterresults} present the results for the Reddit track and the Twitter track, respectively. We show the rank of the submitted systems (best result from their submitted reports) both in terms of the system submissions (out of 14) as well as their rank on the Codalab leaderboard. Note, for a couple of entries we observe a discrepancy between their best reported system(s) and the leaderboard entries. For the sake of fairness, for such cases, we selected the leaderboard entries to present in Table \ref{table:redditresults} and Table \ref{table:twitterresults}. \footnote{Also, for such cases (e.g., abaruah, under the \emph{Approach} column we reported the approach described in the system paper that is not necessarily reflect the scores of Table \ref{table:redditresults}.}   

Also, out of the 14 system descriptions  \emph{duke\_DS} and \emph{AnadKumR} report the performance on the Twitter dataset, only.  For overall results on both tracks, we observe majority of the models outperformed the $LSTM_{attn}$ baseline \cite{ghosh2018clsarcasm}. Almost all the submitted systems have used the transformer-architecture that seems to perform better than RNN-architecture, even without any task-specific fine-tuning.      Although most of the models are similar and perform comparably, we observe a particular system - miroblog - has outperformed the other models in both the tracks by posting an improvement over the 2nd ranked system by more than 7\% F1-score in the Reddit track and by 14\% F1-score in the Twitter track.      

In the following paragraphs, we inspect the performance of the different systems more closely. We discuss a couple of particular aspects.

\paragraph{Context Usage:} One of the prime  
motivating factors for conducting this shared task was to investigate the role of contextual information. We notice the most common approach for integrating context was simply concatenating it with the response text. Novel approaches include :
\begin{enumerate}
\item Taking immediate context as aspect for response in Aspect-based Sentiment Classification architectures (\textbf{taha})
\item CNN-LSTM based summarization of entire context thread (\textbf{salokr})
\item Time-series fusion with proxy labels for context (\textbf{amitjena40})
\item Ensemble of multiple models with different depth of context (\textbf{miroblog})
\item Using explicit separator between context and response when concatenating (\textbf{tanvidadu})
\end{enumerate}

\paragraph{Depth of Context:} Results suggest that beyond three context turns, gains from context information are negligible and may also reduce the performance due to sparsity of long context threads. The depth of context required is dependent on the architecture and CNN-LSTM based summarization of context thread (\textbf{salokr}) was the only approach that effectively used the whole dialogue.

\paragraph{Discrete vs. Embedding Features} The leaderboard was dominated by Transformer based architectures and we saw submissions using BERT or RoBERTa and other variants. Other sentence embedding architectures like Infersent, CNN/LSTM over word embeddings were also used but had middling performances. Discrete features were involved in only two submissions (\textbf{burtenshaw} and \textbf{duke\_DS}) and were the focus of \textbf{burtenshaw} system. %Even though this system had comparatively lesser performance, a thorough comparative analysis of predictions  reveal insights about the sarcasm surface forms. 

\paragraph{Leveraging other datasets} The large difference between the best model (\textbf{miroblog}) and other systems can be attributed to their dataset augmentation strategies. Using just the context thread as a negative example when the context+response is a positive example, is a straight-forward approach for augmentation from labeled dialogues. Their novel contribution lies in leveraging large-scaled unlabelled dialogue threads, showing another use of BERT by using NSP confidence score for assigning pseudo-labels.

\paragraph{Analysis of predictions:} Finally, we conducted an error analysis based on the predictions of the systems. We particularly focused on addressing two questions.  First, we investigate  whether any particular pattern exists in the evaluation instances that are wrongly classified by the majority of the systems. Second, we compare the predictions of the top-performing systems to identify instances correctly classified by the candidate system but missed by the remaining systems. Here, we attempt to recognize specific characteristics that are unique to a model, if any.

Instead of looking at the predictions of all the systems we decided to analyze only the \emph{top-three} submissions in both tracks because of their high performances. We identify 80 instances (30 sarcastic) from the Reddit evaluation dataset and 20 instances (10 sarcastic) from the Twitter evaluation set, respectively, that are missed by all the top-performing systems. Our  interpretation  of  this  finding  is that all these test instances more or less belong to a variety of topics including sarcastic remarks on baseball teams, internet bills, vaccination, etc., that probably do not generalize well during the training. For both Twitter and Reddit, we also found many sarcastic examples that contain common non-sarcastic markers such as laughs (e.g., ``haha''), jokes,  positive-sentiment emoticons (e.g., :)) in terms of Twitter track. We did not find any correlation to context length. Most of the instances contain varied context length, from two to six.

While analyzing the predictions of individual systems we noted that \textbf{miroblog} correctly identifies  the most number of predictions for both the tracks. In fact, miroblog has successfully predicted over two hundred examples (with almost equal distribution of sarcastic and non-sarcastic instances) in comparison to the second-ranked and third-ranked systems for both tracks. As stated earlier, this can be attributed to their data augmentation strategies that have assisted miroblog's models to generalize best. However, we still notice that instances with subtle humor or positive sentiment are missed by the best-performing models even if they are pre-trained on a very large-scale corpora. We foresee models that are able to detect subtle humor or witty wordplay will perform even better in a sarcasm detection task.

% The most obvious take-home message from all the submissions is the overwhelming use of the transformer-based architectures, such as BERT and RoBERTa and other variants. The fact that all but one of the participating teams for the shared task experimented with this architecture testifies to the increasing popularity of applying such approach.  

\section{Conclusion} \label{section:conclusion}

This paper summarizes the results of the shared task on sarcasm detection using conversation from two social media platforms (Reddit and Twitter), organized as part of the 2nd Workshop on the Figurative Language Processing at ACL 2020. This shared task aimed to investigate the role of conversation context for sarcasm detection. The goal was to understand how much conversation context is needed or helpful for sarcasm detection. For Reddit, the training data was sampled from the standard corpus from \newcite{khodak2017large} whereas we curated a new evaluation dataset. For Twitter, both the training and the test datasets are new and collected using standard hashtags. We received 655 submissions (from 39 unique participants) and 1070 submissions (from 38 unique participants) for Reddit and Twitter tracks, respectively. We provided  brief  descriptions  of  each of the  participating  systems who submitted a shared task paper (14 systems).

%Until the last year, the trend in sarcasm detection research was the use of architectures, such as the Attention-based RNN models. 
%From this year's submissions, it is  clear that transformer-based architectures have replaced the RNNs. 
We notice that almost every submitted system have used transformer-based architectures, such as BERT and RoBERTa and other variants, emphasizing the increasing popularity of using pre-trained language models for various classification tasks. The best systems, however, have employed a clever mix of ensemble techniques and/or data augmentation setups, which seem to be a promising direction for future work. We hope that some of the teams will make their implementations publicly available, which would facilitate further research on improving performance on the sarcasm detection task.

\bibliography{acl2020}
\bibliographystyle{acl_natbib}

\end{document}